\documentclass[11pt]{article}
\usepackage{coling2018}
\usepackage{times}
\usepackage{url}
\usepackage{latexsym}
\usepackage{graphicx}
\usepackage{multicol}
\usepackage{array}

\usepackage{floatrow}
\newfloatcommand{capbtabbox}{table}[][\FBwidth]

\usepackage{titlesec}

\titlespacing*{\section} {0pt}{1.0ex plus 1ex minus .2ex}{1.3ex plus .2ex}

\title{Political Advertising Dataset: \\ the use case of the Polish 2020 Presidential Elections}

\author{Łukasz Augustyniak$^1$ \and Krzysztof Rajda$^1$ \and Tomasz Kajdanowicz$^1$ \and Michał Bernaczyk$^2$ \\
    $^1$ Wrocław University of Science and Technology, Department of Computational Intelligence \\ 
    $^2$ Wrocław University, Faculty of Law, Administration and Economics \\
    Wrocław, Poland \\
  {\tt lukasz.augustyniak@pwr.edu.pl}    {\tt krzysztof.rajda@pwr.edu.pl} \\
  {\tt tomasz.kajdanowicz@pwr.edu.pl}    {\tt michal.bernaczyk@uwr.edu.pl}
  }

\date{}

\begin{document}
\maketitle
\begin{abstract}
  Political campaigns are full of political ads posted by candidates on social media. Political advertisements constitute a basic form of campaigning, subjected to various social requirements. We present the first publicly open dataset for detecting specific text chunks and categories of political advertising in the Polish language. It contains 1,705 human-annotated tweets tagged with nine categories, which constitute campaigning under Polish electoral law. We achieved a 0.65 inter-annotator agreement (Cohen's kappa score). An additional annotator resolved the mismatches between the first two annotators improving the consistency and complexity of the annotation process. We used the newly created dataset to train a well established neural tagger (achieving a~70\% percent points F1 score). We also present a possible direction of use cases for such datasets and models with an initial analysis of the Polish 2020 Presidential Elections on Twitter. 
\end{abstract}

\section{Introduction}
\label{sec:intro}

The emergence of social media has changed how political campaigns take place around the world~\cite{Kearney2013}. Political actors (parties, action committees, candidates) utilize social media platforms, Twitter, Facebook, or Instagram, to communicate with and engage voters~\cite{Skogerb2015}. Hence, researchers must analyze these campaigns for several reasons, including enforcement of the laws on campaign contribution limits, implementation of freedom and fairness in electoral campaigns or protection against slander, hate-speech, or foreign interference. Unlike U.S. federal or state law, European jurisdictions present a rather lukewarm attitude to unrestrained online campaigning. Freedom of expression in Europe has more limits~\cite{Rosenfeld2003HateAnalysis}, and that can be seen in various electoral codes in Europe (e.g. in France, Poland) or statutes imposing mandatory systems of notice-and-takedown (e.g. the German Network Enforcement Act, alias the Facebook Act). In Poland, \textit{agitation} (an act of campaigning) may commonly be designated 'political advertisement', corporate jargon originating from such as Twitter's or Facebook's terms of service. Primarily, however, it has a normative definition in article 105 of the Electoral Code. It covers any committees' or voters' public acts of inducement or encouragement to vote for a candidate or in a certain way, regardless of form. Election promises may appear in such activities, but do not constitute a necessary component. A verbal expression on Twitter falls into this category. There exist some natural language resources for the analysis of political content in social media. These include collections related to elections in countries such as Spain~\cite{Taule2018a}, France~\cite{Lai2019b}, and Italy~\cite{Lai2018}. \newcite{Vamvas2020X-Stance:Detection} created an X-stance dataset that consists of German, French, and Italian text, allowing for a cross-lingual evaluation of stance detection. While the datasets on political campaigning are fundamental for studies on social media manipulation~\cite{Aral2019ProtectingManipulation}, there are no Polish language datasets related to either political advertising or stance detection problems. We want to fill this gap and expand natural language resources for the analysis of political content in the Polish language. 
Our contributions are as follows: (1) a novel, publicly open dataset for detecting specific text chunks and categories of political advertising in the Polish language, (2) a publicly available neural-based model for tagging social media content with political advertising that achieves a 70\% F1 score, and (3) an initial analysis of the political advertising during the Polish 2020 Presidential Election campaign.

\section{Political Advertising Dataset Annotation}
\label{sec:dataset}

We created nine categories of political advertising (incl. election promises) based on manually extracted claims of candidates (see sample of examples in Table \ref{tab:examples}) and a taxonomy proposed by \newcite{Vamvas2020X-Stance:Detection}. We gathered political advertising topics (incl. election promises) provided by presidential candidates and committee websites, Facebook fanpages, and other websites (such as news agencies). We tried to fit them into the \newcite{Vamvas2020X-Stance:Detection} taxonomy. However, we spotted that some categories should be corrected or omitted, such as the economy. It was particularly hard to divide whether political advertising should fall into the category of welfare or economy. The final version of categories was created after a couple of iterations with our annotator team, and we chose categories for which we got a consensus and repeatable annotations. By repeatable, we understand that the same or another annotator will consistently choose the same categories for repeated examples or very similar examples. To the best of our knowledge, there do not exist any commonly used political advertising categories. Moreover, we are aware that they can evolve in future election campaigns. We shall be ready to update the dataset according to political advertising types, categories, and any concept drift in the data. 

We extracted all Polish tweets related to the election based on appearances in the tweet of (1) specific hashtags such as \emph{wybory} (\emph{elections}), \emph{wybory2020} (\emph{elections2020}), \emph{wyboryprezydenckie} (\emph{presidentialelections}), \emph{wyboryprezydenckie2020} (\emph{presidentialelections2020}), and (2) unigram and bigram collocations\footnote{Collocation extracted using NLTK library (\url{https://www.nltk.org/}).} generated using examples of political advertising from all categories (as in Table~\ref{tab:examples}). We name these sets of hashtags and collocations together as \emph{search keywords or search terms}. We gathered almost 220,000 tweets covering approximately ten weeks, between February 5, 2020 and April 11, 2020 using the search terms mentioned. We assigned sentiment orientation for each tweet using an external Brand24 API\footnote{\url{https://brand24.com/}}. Then we sampled tweets using a two stage procedure: (1) we divided all tweets into three sets based on sentiment orientation (positive, neutral, negative) to have representations of various attitudes, and (2) for each sentiment category we randomly selected $log_{2}(|E_{k}|)$ examples, where $E$ is a set of all tweets for a particular \emph{search keyword} $k$. We used almost 200 different \emph{search keywords} and finally reached 1,705 tweets for the annotation process.
The dataset was annotated by two expert native speaker annotators (linguists by training). The annotation procedure was similar to Named Entity Tagging or Part-of-Speech Tagging, so the annotators marked each chunk of non-overlapping text that represents a~particular category. They could annotate a single word or multi-word text spans. Table \ref{tab:examples} presents an example of political advertising with corresponding categories. These kinds of examples (they could be named as seed examples) were presented to the annotators as a starting point for annotating tweets. However, the annotators could also mark other chunks of text related to political advertising when the chunk is semantically similar to examples or it clearly contains examples of a political category but not present in the seed set. We achieved a 0.48 Cohen's kappa score for exact matches of annotations (even a~one-word mismatch was treated as an error) and a 0.65 kappa coefficient counting partial matches such as \emph{reduce coil usage} and \emph{reduce coil} as correct agreement. We disambiguated and improved the annotation via an additional pass by the third annotator. He resolved the mismatches between the first two annotators and made the dataset more consistent and comprehensive. According to \newcite{McHugh2012}, the 0.65 kappa coefficient lies between a moderate and strong level of agreement. We must remember that we are annotating social media content. Hence there will exist a lot of ambiguity, slang language, and very short or chunked sentences without context that could influence the outcome.
Then, we trained a~Convolutional Neural Network model using a spaCy Named Entity classifier~\cite{Mather}, achieving a 70\% F1 score for 5-fold cross-validation. We used fastText vectors for Polish~\cite{Grave2019a} and default spaCy model hyperparameters. The dataset and trained model are publicly available in the GitHub repository \footnote{\url{https://github.com/laugustyniak/misinformation}}. Table \ref{tab:scores} presents per category precision, recall, F1 score, as well as the number of examples for each category in the whole dataset. As we can see, there are 2507 spans annotated. Interestingly, 631 tweets have been annotated with two or more spans. Finally, 235 tweets do not contain any annotation span, and they represent 13.8\% of the whole dataset.

\begin{table}[]

\def\arraystretch{0.65}
\begin{tabular}{|m{1.1cm}|m{1.2cm}|m{1.4cm}|m{1.3cm}|m{1.5cm}|m{1.5cm}|m{1.5cm}|m{1.4cm}|m{1.2cm}|}

%\hline

\scriptsize \centering \textbf{Healthcare} & 
\scriptsize \centering \textbf{Welfare} & 
\scriptsize \centering \textbf{Education} & 
\scriptsize \centering \textbf{Immigration} & 
\scriptsize \centering \textbf{Infrastructure and Environment} & 
\scriptsize \centering \textbf{Defense and Security} & 
\scriptsize \centering \textbf{Foreign Policy} & 
\scriptsize \centering \textbf{Society} & 
\scriptsize \centering \textbf{Political and Legal System} \tabularnewline

\hline

\scriptsize \centering drug reform, national oncology strategy & 
\scriptsize \centering tax-free allowance, 500+ programme, minimum wage &
\scriptsize \centering e-learning, remote learning, dentist in every school & 
\scriptsize \centering immigrant, refugees &
\scriptsize \centering climate change, reduce coil usage, water crisis &
\scriptsize \centering national security, alliance with the USA, NATO alliance & 
\scriptsize \centering lack of foreign policy, EU collaboration, Weimar Triangle & 
\scriptsize \centering death penalty, separation of church and state, LGBT & 
\scriptsize \centering presidential veto, independence of the courts \tabularnewline
\hline

\end{tabular}
\caption{Political advertising -- categories with examples.\label{tab:examples}}
\end{table}

\section{Polish 2020 Presidential Election - Use Case}
\label{sec:use_case}
\begin{figure}[t]
\begin{floatrow}
    \capbtabbox{
        \begin{tabular}{|c|c|c|c|c|}
        \label{tab:scores}
%        \hline
        \small Category & P & R & F1 & \#
        \\ \hline
        \small Healthcare & 0.75 & 0.72 & 0.74 & 586 \\
        \small Welfare & 0.76 & 0.60 & 0.67 & 526\\
        \small Defense & 0.92 & 0.71 & 0.80 & 57\\
        \small Legal & 0.79 & 0.59 & 0.68 & 352\\
        \small Education & 0.81 & 0.57 & 0.67 & 163\\
        \small Infrastructure & 0.77 & 0.59 & 0.67 & 284\\
        \small Society & 0.82 & 0.67 & 0.74 & 386 \\
        \small Foreign Policy & 0.86 & 0.43 & 0.57 & 93\\
        \small Immigration & 0.89 & 0.67 & 0.76 & 60\\
        \hline
        \small  & \textbf{0.82} & \textbf{0.62} & \textbf{0.70} & \textbf{2507} \\
        \hline
        \end{tabular}
        
    }{
        \caption{Precision, Recall, and F1 score for each of the promises categories. The last column presents the number of examples in the dataset.}
    }

  \hfill
  \begin{minipage}[t]{0.55\textwidth}
          \ffigbox
          {
                % \vspace*{-1.5in}
                % \setlength{\belowcaptionskip}{-10pt}
                % \centering
                \label{fig:sentiment}
                \includegraphics[width=\linewidth]{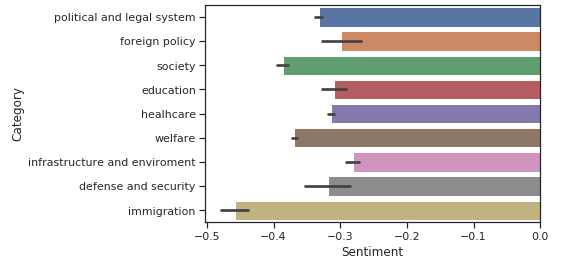}
            }
            {
                \caption{Sentiment orientation for each of political advertising category. The bars mean standard deviations. }
            }
      \end{minipage}

\end{floatrow}
\end{figure}

We present the analysis of 250,000 tweets related to the Polish 2020 Presidential Elections gathered between February 2, 2020 and April 23, 2020. The data acquisition and sentiment assignment procedures were similar to those described in Section \ref{sec:dataset}. The dataset and model we propose enabled us to analyze sentiment polarity across all election promise categories. Figure \ref{fig:sentiment} shows the overall average sentiment categories. The sentiment has been assigned on a -1 (negative) to 1 (positive) scale. None of the categories were positive on average; hence for readability we show only the zoomed part of the graph with a~scale from -0.5 (moderate negative) to 0 (neutral) sentiment. All categories contain a~much more negative attitude on the scale of an absolute sentiment analysis score. As we can imagine, most tweets gathered by us are from potential voters, and there are many more negative than positive messages. Most of the sentiment analysis tools available right now perform only general sentiment detection, saying only how many positive or negative tweets they have identified and analyzed. However, our dataset and model enable us to go deeper into the analysis of attitudes towards particular political advertising categories or even more granular towards specific election promises.

%\vfill

\section{Conclusions and Future Work}
\label{sec:conclusions}
A new dataset and model enables us to analyze the Polish political scene, counter political misinformation in social media, and evaluate the political advertising of candidates. We plan to work on more datasets and models to fight fake news, classify political agitation content, and widen natural language solutions in regards to elections and political content in Polish social media. The dataset annotation will be very challenging due to many potential concept drifts between different election types such as presidential, parliamentary, European Union, and others. 
% We want to utilize this analysis of the 2020 Polish Presidential Election and future campaigns. 
We use  political advertising model to generate presidential candidates' vector representations. We can compare candidates with each other and say who is similar to whom.

\section*{Acknowledgment}
\small The work was partially supported by the National Science Centre, Poland grant No. 2016/21/N/ST6/02366 and by the Faculty of Computer Science and Management, Wrocław University of Science and Technology statutory funds.

\nocite{Tumasjan2010, Burnap2016, Heredia2017, Prasetyo2015, Bayerl2011, Kearney2013, Gorwa2017, Bahuleyan2018a, Kochkina2018b, Ceron2016, Cohen2013, Gayo-Avello2012, Yaqub2017, Best2015, Isaak2018, Rossi2020, Golovchenko2020, Kearney2013, Bernaczyk2020ZnaczeniePolitycznego}
\bibliographystyle{acl}
\bibliography{acl}

\end{document}